\documentclass[letterpaper]{article} % DO NOT CHANGE THIS
\usepackage{aaai25}  % DO NOT CHANGE THIS
\usepackage{times}  % DO NOT CHANGE THIS
\usepackage{helvet}  % DO NOT CHANGE THIS
\usepackage{courier}  % DO NOT CHANGE THIS
\usepackage[hyphens]{url}  % DO NOT CHANGE THIS
\usepackage{graphicx} % DO NOT CHANGE THIS
\usepackage{natbib}  % DO NOT CHANGE THIS AND DO NOT ADD ANY OPTIONS TO IT
\usepackage{caption} % DO NOT CHANGE THIS AND DO NOT ADD ANY OPTIONS TO IT
\usepackage{algorithm}
\usepackage{algorithmic}
\usepackage{amsmath}
\usepackage{listings}
\usepackage{graphicx}
\usepackage{tcolorbox}
\usepackage{booktabs}

\usepackage{amsmath}
\usepackage{amsthm}
\usepackage{amsfonts}
\usepackage{colortbl}
\usepackage{multirow}
\usepackage{adjustbox}
\usepackage{array}

\usepackage{xcolor}
\usepackage{tcolorbox}

\definecolor{lightyellow}{RGB}{255,255,204}
\definecolor{lightgreen}{RGB}{235,255,235}
\definecolor{lightred}{RGB}{255,204,203}
\definecolor{lightblue}{RGB}{202,225,255} % Definition for light blue color
\definecolor{validitygreen}{RGB}{0,255,0}

\usepackage{tcolorbox}
\usepackage{listings}
\usepackage{color}

\definecolor{mygreen}{rgb}{0,0.6,0}
\definecolor{mygray}{rgb}{0.5,0.5,0.5}
\definecolor{mymauve}{rgb}{0.58,0,0.82}

\usepackage{newfloat}
\usepackage{listings}
\DeclareCaptionStyle{ruled}{labelfont=normalfont,labelsep=colon,strut=off} % DO NOT CHANGE THIS
\floatstyle{ruled}
\newfloat{listing}{tb}{lst}{}
\floatname{listing}{Listing}
\usepackage{listings}
\usepackage{color}

\definecolor{myblue}{RGB}{0, 0, 255}
\definecolor{myred}{RGB}{220, 20, 60}
\definecolor{mypurple}{RGB}{128, 0, 128}
\definecolor{myblack}{RGB}{0, 0, 0}

\title{Knowledge Graph Modeling-Driven Large Language Model Operating System (LLM OS) for Task Automation in Process Engineering Problem-Solving}
\author{
    Sakhinana Sagar Srinivas\textsuperscript{\rm 1}\thanks{corresponding author.},
    Vijay Sri Vaikunth\textsuperscript{\rm 2},
    Venkataramana Runkana\textsuperscript{\rm 1}\\
}
\affiliations{
    \textsuperscript{\rm 1} TCS Research, \textsuperscript{\rm 2} IIT–Palakkad \\
    \texttt{sagar.sakhinana@tcs.com}, \texttt{112101060@smail.iitpkd.ac.in}, \texttt{venkat.runkana@tcs.com}\\
}

\usepackage{bibentry}
\begin{document}

\maketitle

\begin{abstract}
\vspace{-2mm}
We present the Process Engineering Operations Assistant (PEOA), an AI-driven framework designed to solve complex problems in the chemical and process industries. The framework employs a modular architecture orchestrated by a meta-agent, which serves as the central coordinator, managing an action generator and instruction-tuned small-scale language models (expert models). The action generator decomposes complex problems into sub-tasks and identifies suitable expert models to execute each, delivering precise solutions for multi-step problem-solving. Key techniques include advanced knowledge modeling using property graphs for improved information retrieval, facilitating more accurate and contextually relevant solutions. Additionally, the framework utilizes a teacher-student transfer-learning approach with GPT-4 (Omni) to fine-tune the action generator and expert models for domain adaptation, alongside an iterative problem-solving mechanism with sophisticated error handling. Custom datasets were developed to evaluate the framework against leading proprietary language models on various engineering tasks. The results demonstrate the framework's effectiveness in automating calculations, accelerating prototyping, and providing AI-augmented decision support for industrial processes, marking a significant advancement in process engineering capabilities.
\vspace{-3mm}
\end{abstract}

\vspace{-3mm}
\section{Introduction}
\begin{figure}[!ht]
\vspace{-2mm}
\centering
\resizebox{1.1\linewidth}{!}{ 
\hspace*{0mm}\includegraphics[keepaspectratio,trim=0.0cm 0cm 0cm 0.35cm,clip]{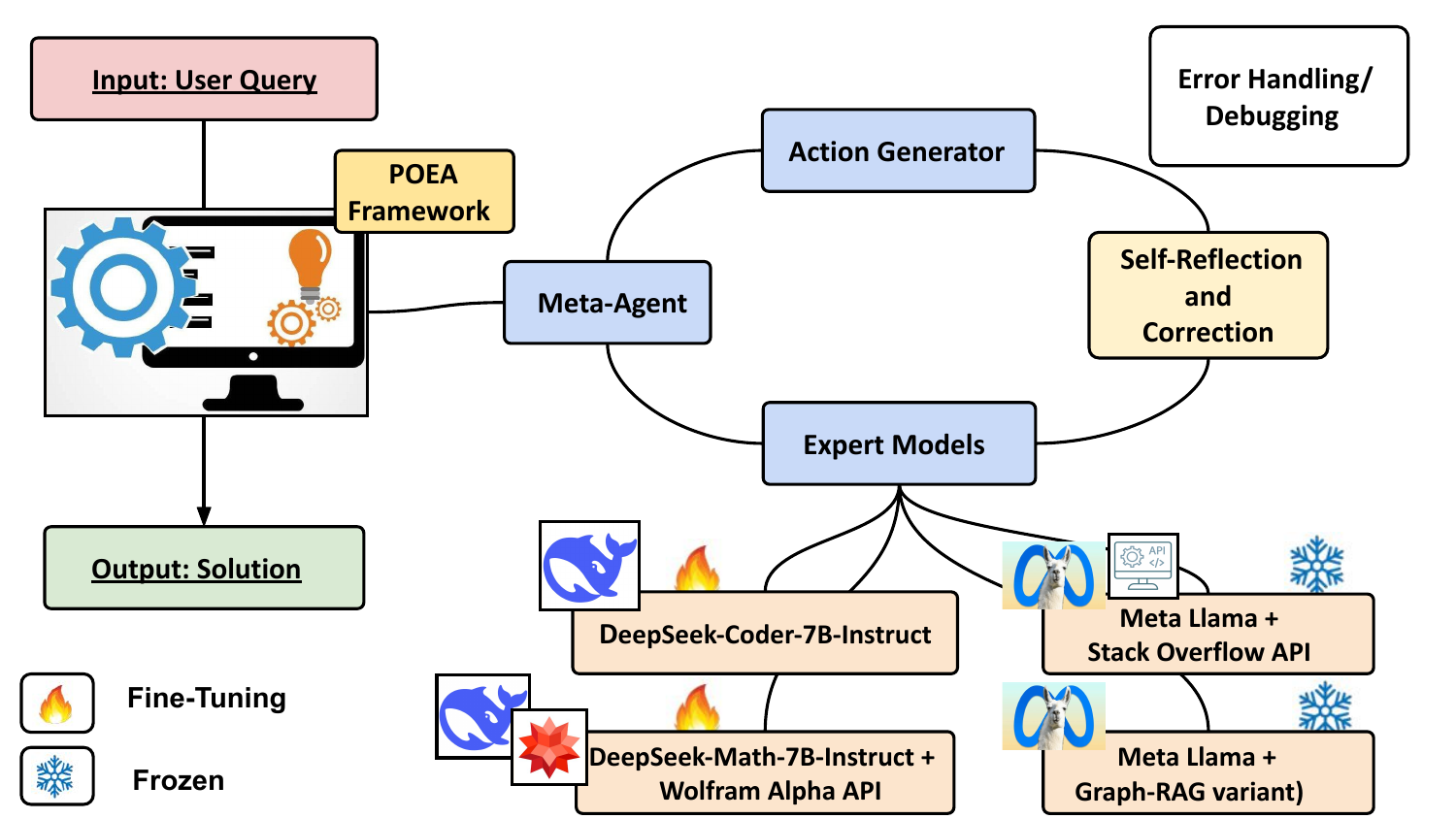} % left, bottom, right, top
}
\vspace{-5mm}
\caption{The figure shows the architecture of the \texttt{PEOA} framework. It illustrates the key components and data flow of the framework, including the central meta-agent, action generator, and expert models. The framework processes user queries, selects specialized tools as expert models for sub-task solving, and employs error-handling mechanisms to generate solutions for complex chemical and process engineering problems. The interconnected nature of the components highlights the framework's ability to decompose tasks, select appropriate tools, and iteratively refine solutions through a sophisticated orchestration process.
}
\label{fig:figure1}
\vspace{-3mm}
\end{figure}

In recent years, significant advancements have been made in retrieval-augmented generation techniques (RAG), which combine the capabilities of large language models (LLMs) with external knowledge sources to enhance information retrieval and question-answering tasks. However, while traditional RAG techniques excel at localized information retrieval, they struggle with global questions requiring a holistic understanding of knowledge bases. Recently, there has been a surge of interest in the Graph RAG approach \cite{sciphi2024triplex, Edge2024GRAG, Hu2023GRAG}, which integrates the strengths of property graph-based knowledge modeling from unstructured data and graph-based indexing with the retrieval and generation capabilities of LLMs. By leveraging these strengths, the Graph RAG approach aims to overcome the limitations of traditional RAG techniques. The combined market capitalization of the Oil and Gas, Semiconductor, Fast-moving consumer goods (FMCG), Pharmaceuticals, Automobile, Aviation, and Energy sectors amounts to approximately \$20 trillion USD. These industrial/manufacturing sectors involve complex chemical and process engineering challenges from a broader perspective. By enhancing problem-solving capabilities in these major industries with Graph RAG approaches for both process knowledge graph modeling and retrieval for question-answering (Q\&A) tasks, we have the potential to contribute to economic growth, technological advancement, and improved competitiveness on a global scale. The rapidly evolving landscape of chemical and process engineering presents numerous complex challenges that necessitate innovative solutions for design, optimization, and troubleshooting. To address these challenges, we present the \textbf{Process Engineering Operations Assistant (PEOA)} framework—a modular, AI-driven Large Language Model Operating System (LLM OS) designed to tackle intricate problems in the chemical and process industry by automating key steps in the problem-solving process. The framework architecture revolves around a central orchestrator, or meta-agent, which coordinates the framework's various components. The meta-agent works in tandem with an action generator, which plays a key role in breaking down complex problems into manageable sub-tasks and identifying the most appropriate tools (or expert models) for each step in solving the sub-tasks. To execute these sub-tasks with high precision, the action generator employs a collection of expert models, each specialized in different aspects. In essence, the framework utilizes a two-stage pipeline that iteratively decomposes complex problems into manageable sub-tasks, selects and chains together suitable tools, and executes solutions. The (subject-matter) expert models include small-scale language models (SLMs) for code generation, mathematical reasoning, and structured information retrieval from property graphs, enabling the framework to leverage external knowledge and solve diverse problems by decomposing, executing, and refining multi-step problem-solving trajectories. The framework incorporates an advanced error-handling mechanism. When a runtime error occurs, it uses a reflection procedure to identify the faulty step and associated tool (expert model). An expert model then generates a revised solution, considering both the immediate error and the broader problem context. This procedure iterates until a successful solution is achieved or a predefined limit is reached. The debugging mechanism functions in two phases: error identification and solution revision. It allows the framework to dynamically adapt its problem-solving strategy, refine solutions iteratively, and tackle increasingly complex tasks that may require multiple rounds of adjustment, thereby improving its robustness and effectiveness in real-world engineering scenarios. The proposed framework addresses the limitations of current language models and problem-solving approaches in the industry, which are hindered by a lack of domain-specific knowledge and expertise, an inability to integrate diverse tools and data sources, and a limited capacity for complex, multi-step reasoning. These limitations result in inefficient and time-consuming problem-solving workflows that impede innovation and progress in the chemical and process industry. The proposed framework serves as a decision support tool, enabling process engineers to focus on high-level decision-making and innovation, accelerate design cycles through rapid prototyping and testing, and optimize chemical processes to enhance yield, efficiency, and safety. Figure \ref{fig:figure1} illustrates the framework.

A key challenge is the lack of tool-integrated solutions for the chemical and process domain. To address this, we use a teacher-student transfer-learning approach with GPT-4 (Omni) as the teacher model to create tool-integrated solution trajectories. These serve as synthetic datasets for customizing the \texttt{PEOA} framework, generating detailed, step-by-step solutions that facilitate the transfer of advanced problem-solving capabilities to the student model. At its core, the framework utilizes a modular architecture that combines instruction-tuned small-scale language models (expert models) with graph retrieval-augmented code generation capabilities, leveraging knowledge graph databases for multi-hop reasoning and improved factual accuracy. For graph retrieval, we use an advanced knowledge modeling technique that parses complex documents (scholarly articles), constructs semantic knowledge graphs (i.e., transforming these documents into structured, searchable graphs), and indexes them for efficient information retrieval. Instruction-tuning small-scale language models (SLMs) like expert models is crucial because they often lack extensive pre-trained knowledge and specialized problem-solving skills needed for complex domain-specific tasks in chemical and process engineering. Unlike proprietary large-scale models such as GPT-4 (Omni), which have more comprehensive pre-trained knowledge, expert models require adaptation to effectively utilize external information, such as language models with vector similarity search on knowledge graphs, resulting in more accurate and efficient solutions. By using instruction tuning with Graph Retrieval-Augmented Code Generation (GRACG), the framework can generate structured, multi-step solution trajectories that systematically solve complex tasks. To evaluate the proposed framework, we developed custom datasets focused on mathematical modeling, computational methods, and chemical and process engineering. We conducted extensive experiments comparing the framework's performance to leading proprietary LLMs on a range of complex engineering tasks. Our work is the first step in significantly enhancing the capabilities of process engineers by automating routine calculations, accelerating prototyping and optimization, and providing AI-augmented decision support for complex industrial processes. The framework can manage the lifecycle of SLMs (expert models), including fine-tuning, monitoring, and updating these models. These tools enable the maintenance of the framework's accuracy and relevance over time, ensuring optimal performance and decision support for complex industrial processes. In summary, the \texttt{PEOA} framework represents a significant advancement in automating complex problem-solving in chemical and process engineering and offers a powerful solution for optimizing processes, accelerating innovation, and supporting high-level decision-making in this challenging field.

\vspace{-2mm} 
\section{Related Work}
\vspace{-1mm}
Large Language Models (LLMs) have demonstrated notable capabilities in various reasoning tasks, including those involving graph-structured data. However, despite their success, LLMs often face challenges with factual accuracy due to limitations in their training data and a lack of real-time knowledge integration \cite{hu2023a}. To address these issues, Retrieval-Augmented Generation (RAG) has been developed, enhancing LLMs by integrating external data retrieval into the generative process, which improves the relevance and accuracy of responses \cite{lewis2020}. Traditional RAG approaches, however, focus mainly on text-based entity retrieval and often overlook the structural intricacies of graph data, which are critical for tasks requiring multi-hop reasoning and context preservation across documents to answer global queries \cite{yasunaga2021}. For example, conventional RAG methods split text into chunks, map these into a vector space, and measure similarity with the query vector but fail to capture the topological information inherent in graphs \cite{velivckovic2018}. The integration of graphs with LLMs and RAG is an emerging research area. Previous work has explored using LLMs for knowledge graph creation \cite{trajanoska2023enhancing}, completion \cite{yao2023exploring}, and causal graph extraction \cite{ban2023query, zhang2024causal}. Advanced RAG methods leverage graph structures as knowledge indexes \cite{baek2023knowledge}, use subsets of graphs for answering queries \cite{he2024g, zhang2023graph}, and ground narrative outputs in subgraphs \cite{kang2023knowledge}. Recently, a Graph RAG approach \cite{Edge2024GRAG} has been introduced that utilizes LLMs to construct knowledge graphs and employs graph modularity and community detection to generate comprehensive, diverse query-focused summaries for Q\&A tasks. Additionally, Triplex \cite{sciphi2024triplex, sciphi2024huggingface}, an advanced language model for efficient knowledge graph construction, extracts subject-predicate-object triplets from unstructured data, offering significant cost reductions and improved performance compared to traditional methods and general-purpose models like GPT-4. Several open-source libraries now support graph databases for RAG applications. For instance, LangChain \cite{langchain2024} and LlamaIndex \cite{llamaindex2024} facilitate graph-based RAG applications with integration into Neo4j \cite{neo4j2024} and NebulaGraph \cite{nebulagraph2024}. These advancements enhance the performance and scalability of Graph RAG systems by structuring information in a modular and hierarchical manner. At its core, the \texttt{PEOA} framework utilizes advanced knowledge graph construction and Graph Retrieval-Augmented Code Generation (GRACG) to tackle complex chemical and process engineering challenges. By transforming unstructured data into structured, context-rich graphs, the framework enables efficient, context-aware querying while preserving relationships and integrating diverse data types. This approach improves problem-solving capabilities by leveraging domain-specific tools (e.g., expert models) and creating structured solution trajectories, thereby enhancing accuracy and efficiency. The combination of knowledge modeling through property graphs and graph-retrieval augmentation allows the framework to deliver precise, systematic solutions, streamline workflows, and automate complex engineering tasks, accelerating design cycles and supporting high-level decision-making in the process industry.

\vspace{-3mm} 
\section{Proposed Method}
\vspace{0mm}
We aim to address the complex challenges faced by chemical and process engineers in designing, optimizing, and troubleshooting industrial processes. To this end, we are developing the \textbf{Process Engineering Operations Assistant} (\texttt{PEOA}), a task automation framework conceptualized as a \textbf{Large Language Model Operating System (LLM OS)}. This modular framework combines AI-driven capabilities with computational tools to streamline problem-solving in chemical and process engineering. At its core, the \texttt{PEOA} framework leverages the LLM OS to manage and orchestrate foundational language models, automating key steps in the problem-solving process. The objective is to allow process engineers to focus on high-level decisions and innovation while accelerating design cycles through faster prototyping and testing and optimizing chemical processes to identify optimal conditions that maximize yield, efficiency, and safety. Small-scale language models for code (SLMs) such as Google Code Gemma \cite{team2024codegemma} and Meta Code Llama \cite{codellama} often lack extensive pre-trained knowledge related to domain-specific tasks, such as specialized mathematical reasoning and problem-solving skills for the chemical and process industry, compared to proprietary large-scale models (LLMs) like GPT-4 (Omni) \cite{achiam2023gpt}. Additionally, SLMs are not designed to effectively incorporate and utilize external knowledge from various domain-specific tools (e.g., vector similarity search on knowledge graph databases of code repositories/documentation, or retrieval-augmented generation with Stack Overflow APIs) for more accurate and efficient problem-solving beyond their pre-trained knowledge \cite{zhang2024raft}. These limitations hinder the performance of SLMs in specialized domains. Instruction-tuning SLMs to access external information offers a promising solution, improving their use of relevant background knowledge for more accurate outputs. We utilize \textbf{Instruction Tuning} with \textbf{Graph Retrieval-Augmented Code Generation (GRACG)}, allowing the proposed framework utilizing SLMs combined with the ReAct (Reason + Act) \cite{yao2022react} prompting technique, to generate `solution trajectories'—structured, step-by-step problem-solving sequences that break down complex tasks, integrate various tools, and produce coherent solutions. Unlike traditional RAG, which relies on linear text retrieval, GRACG techniques utilize knowledge graph databases that preserve graph topology. Graph-based representation of relationships and hierarchies between entities and concepts is more effective than flat text, providing richer contextual information, enhancing multi-hop reasoning, and reducing hallucinations. The solution trajectory consists of multiple steps, executed sequentially to systematically and incrementally solve complex chemical and process industry problems. Each step in a trajectory includes a high-level step description (a subtask), a specific tool to use from a predefined set, and the tool-executed output reformulated in natural language. While this approach shows promise, a significant challenge remains in its implementation. There isn't a pre-existing tool-integrated problem-solving solution trajectory (curated dataset) relevant to the domain that illustrates the comprehensive process of solving complex, multi-step reasoning tasks step-by-step through the integration of various tools. Such a trajectory would provide a structured approach for instruction-tuning SLMs by enhancing domain-specific knowledge and computational tool usage to generate code for solving process engineering calculations. To overcome this limitation, we utilize a \textbf{teacher-student learning} paradigm \cite{kim2024husky} for adapting SLMs to domain-specific tasks with similar performance to proprietary LLMs. It involves a foundational LLM, such as GPT-4 (Omni), serving as a robust teacher (subject-matter expert) to generate high-quality, tool-integrated solution trajectories that serve as synthetic, instruction-tuning datasets demonstrating effective problem-solving strategies. The machine-generated datasets are used to develop a robust and customizable student model—\texttt{PEOA}—for solving process engineering calculations. The teacher model is prompted with few-shot examples to generate step-by-step solutions that involve calling specific tools to solve domain-specific tasks. Each solution trajectory consists of a sequence of steps (i.e., subtasks to perform), with corresponding tools to be used at each step, and outputs (i.e., the result of executing the tool on the given step) reformulated into natural language. Our method efficiently transfers knowledge from the large teacher model by distilling its advanced mathematical reasoning and problem-solving capabilities to a smaller student model. The student model learns effective strategies for performing complex multi-step reasoning, breaking down complex tasks into smaller, more manageable steps, integrating diverse tools, and producing coherent step-by-step solutions (tool-specific outputs). Tools are specialized components or services that enhance the capabilities of language models, enabling them to handle complex and diverse tasks. These include code generators for creating executable snippets, math problem solvers for mathematical reasoning, and vector-search retrieval on knowledge graph databases for structured information access. By integrating these tools, language models can expand their problem-solving abilities. Integrating external tools with automatic tool chaining \cite{shi2024chain} allows SLMs to execute tasks beyond their pre-trained knowledge, augmenting their problem-solving abilities. Tool learning involves four stages: it begins with \textbf{task planning}, where the SLM analyzes a user's query and decomposes it into sub-tasks with tuning-free methods, like few-shot prompting with ReACT techniques. Next, in \textbf{tool selection}, the SLM identifies the most appropriate tools for each sub-task. During \textbf{tool calling}, the SLM extracts and formats the necessary parameters from the user's query to invoke the selected tools. Finally, in \textbf{response generation}, the SLM synthesizes the tool outputs with its own pre-trained knowledge to provide a comprehensive and coherent response. Tool learning can follow two paradigms: \textbf{one-step task solving}, where SLMs plan sub-tasks upfront and generate responses without adjusting for errors, and \textbf{iterative task solving}, where SLMs interact with tools iteratively, correcting tool outputs based on feedback. In this work, we use iterative task solving to enable SLMs to handle complex queries more effectively by leveraging external tool chaining. Given a natural language query \( Q \), we begin by decomposing it into smaller, manageable sub-tasks. Let \( \mathcal{S} = \{s_1, s_2, \ldots, s_n\} \) be the set of sub-tasks derived from \( Q \). The aim is to enable the proposed framework to use a sequence of tools from the set \( \mathcal{T} = \{t_1, t_2, \ldots, t_{|\mathcal{T}|}\} \) to solve the task. For each sub-task \( s_i \), the most appropriate tool \( t_i \) is selected from the set of tools \( \mathcal{T} \). The framework first determines if tool usage is necessary to solve the sub-task. If so, the program chains them together to complete the task. When tools are not required, the framework relies on its internal pre-trained knowledge to solve the task. The tool protocols provide meta-information to understand each tool's purpose and usage. The tool protocols \( \mathcal{D} = \{d_1, d_2, \ldots, d_{|\mathcal{D}|}\} \) consist of documented protocols \( d_i \) corresponding to each tool \( t_i \in \mathcal{T} \). Each protocol \( d_i \in \mathcal{D} \) offers detailed information about its associated tool \( t_i \), including an overview of functionality and use cases, argument requirements specifying necessary inputs, and a response schema outlining expected output structure and type. The detailed tool protocols allow the framework to learn tool usage, understand the input-output schema and capabilities of various tools, and manage data flow dependencies, enabling it to chain together and utilize multiple tools to solve the end-user task. Tool learning is a crucial component of the proposed framework, supporting its core objective of streamlining workflows and automating complex problem-solving tasks in process engineering. The \texttt{PEOA} framework consists of a meta (top-level) agent orchestrating a specialized action generator (\(\mathcal{A}\)) and expert models (tools) (\(\mathcal{M}_{t}\)) modeled by SLMs. The meta agent delegates the input question and the solution history to the action generator to predict the next high-level sub-task and select the appropriate tool needed to solve the sub-task, which expert models then execute precisely, updating the solution state. The framework iterates over a two-stage pipeline to solve multi-step reasoning tasks using various expert models as tools, combining the generation of sub-tasks and tool selection followed by invoking the specialized expert models to efficiently address complex problems. Tool-integrated solution trajectories generated by the teacher model fine-tune the action generator and expert models. The framework employs a diverse set of expert models to execute actions based on the tool chosen by the action generator. These models include \(\mathcal{M}_{c}\) (DeepSeek-Coder-7B-Instruct \cite{guo2024deepseek}) for generating executable code snippets, and \(\mathcal{M}_{m}\), a RAG technique variant that combines the mathematical reasoning of DeepSeek-Math-7B-Instruct \cite{shao2024deepseekmath} with the computational power of Wolfram Alpha's \cite{hindin2010wolfram} API for advanced problem-solving. Additionally, \(\mathcal{M}_{q}\) (Meta Llama or Google Gemma) is used for crafting search queries, translating sub-tasks into understandable formats to retrieve information from web search engines like DuckDuckGo or Stack Overflow APIs, and parsing their outputs. Lastly, \(\mathcal{M}_{KQ}\) (Meta Llama or Google Gemma), a graph-RAG variant, is employed for conducting structured information retrieval through similarity searches from knowledge graph databases of scholarly sources such as numerical libraries/code documentation. The action generator (\(\mathcal{A}\)) is realized with Meta Llama or Google Gemma. Finally, the top-level agent integrates the results, potentially with its own knowledge, to craft a coherent, human-friendly response that provides context, explanations, and insights, allowing the framework to tackle a wide range of complex tasks by leveraging specialized tools as needed. The action generator \(\mathcal{A}\) takes the task instruction \(x\) and the concatenated solution history \(h_{i-1}\) up to the previous step and predicts the next step \(s_i\) and the associated tool \(t_i\) to solve the sub-task, described as follows:

\vspace{-2mm}
\resizebox{0.925\linewidth}{!}{
\hspace{0mm}\begin{minipage}{\linewidth}
\begin{equation}
 \mathcal{A}(I_{a}, x, h_{i-1}) = \mathcal{A}(I_{a}, x, [s_1 \, \| \, o_1 \, \| \, \ldots \, \| \, s_{i-1} \, \| \, o_{i-1}]) \rightarrow [t_i, s_i] \nonumber
\end{equation}
\end{minipage}
}

\vspace{1mm}
where \(h_{i-1}\) is the solution history up to step \(i-1\) and \(I_{a}\) indicates a concise instruction prompt provided to the action generator to predict the next step \(s_i\) and the tool \(t_i\). The step \(s_i\) is the high-level description of the action to be taken at each stage in the tool-integrated solution trajectory. The expert model \(\mathcal{M}_{t_i}\) associated with the tool \(t_i\) generates the output \(o_i\) for the step \(s_i\) as follows:

\vspace{0mm}
\resizebox{0.925\linewidth}{!}{
\hspace{0mm}\begin{minipage}{\linewidth}
\begin{equation}
\mathcal{M}_{t_i}(I_{m}, x, h_{i-1}, s_i) \rightarrow o_i \nonumber
\end{equation}
\end{minipage}
}

\vspace{1mm}
where \(\mathcal{M}_{t_i}\) is the expert model corresponding to the tool \(t_i\), \(o_i\) is the output of the current step, and \(h_i = h_{i-1} \cup (s_i, o_i)\) is the updated solution history including \(s_i\) and \(o_i\). \(I_{m}\) serves as a concise instruction prompt provided to the expert model to generate the output for a given step in the solution trajectory. The output \(o_i\) is generated by executing the tool's action. For example, a code snippet \(c_i\) generated by \(\mathcal{M}_c\) is executed by a code interpreter to produce \(o_i\). The iterative process continues until the action generator \(\mathcal{A}\) identifies the final answer to \(x\) in the solution history \(h\). \(\mathcal{A}\) and \(\mathcal{M}_t\) are trained on tool-integrated solution trajectories generated by a teacher LM (GPT-4 (Omni)). At inference time, the proposed framework uses \(\mathcal{A}\) to predict steps (sub-tasks) and tools, and \(\mathcal{M}_t\) to execute these steps until it finds the final answer. Table \ref{tab:co2_volume_calculation} demonstrates the framework's ability to break down a complex problem into manageable steps, utilize appropriate tools (in this case, code execution), and provide a clear, step-by-step solution. Note: The example was chosen for simplicity and illustration.

\vspace{-0.3cm}
\begin{table}[ht!]
\begin{tcolorbox}[colback=white!5!white, colframe=black!75!black, title=Calculate the volume occupied by 88 lb of CO$_2$ at 15°C and a pressure of 32.2 ft of water., fontupper=\bfseries\small]

\vspace{-0.1cm}
\textbf{Step 1: Convert pressure from ft of water to atm}

\vspace{0.1cm}
\textbf{Tool:} [code]

\vspace{0.1cm}
\textbf{Code:}
\vspace{-0.1cm}
\begin{verbatim}
# Constants
ft_to_atm = 0.0294
pressure_ft = 32.2
# Convert pressure
pressure_atm = pressure_ft * ft_to_atm
print(pressure_atm)
\end{verbatim}

\vspace{0.1cm}
\textbf{Output:} 0.94668

\vspace{0.1cm}
\textbf{Result:} The pressure is 0.94668 atm.

\vspace{0.3cm}
\textbf{Step 2: Use the Ideal Gas Law to calculate the volume}

\vspace{0.1cm}
\textbf{Tool:} [code]

\vspace{0.1cm}
\textbf{Code:}
\vspace{-0.1cm}
\begin{verbatim}
# Constants
R = 0.0821  # L·atm/(K·mol)
T = 288.15  # K
n = 2  # moles
P = 0.94668  # atm
# Ideal Gas Law
V = (n * R * T) / P
print(V)
\end{verbatim}

\vspace{0.1cm}
\textbf{Output:} 49.867

\vspace{0.1cm}
\textbf{Result:} The volume occupied by 88 lb of CO$_2$ is 49.8 L.

\vspace{0.0cm}
\textbf{Final Answer:} The volume occupied by 88 lb of CO$_2$ at 15°C and a pressure of 32.2 ft of water is 49.8 liters.
\vspace{-0.1cm}
\end{tcolorbox}
\vspace{-5mm}
\caption{Example of a tool-integrated solution trajectory for calculating the volume of CO$_2$.}
\label{tab:co2_volume_calculation}
\vspace{-3mm}
\end{table}

The framework employs a sophisticated error-handling (code debugging) \cite{gou2023critic} and adaptive problem-solving mechanism, utilizing a dynamic interplay between an action generator \(\mathcal{A}\) and specialized expert models \(\mathcal{M}_{t}\), which work in tandem to decompose, execute, and refine multi-step problem-solving trajectories. When encountering a runtime error, the action generator \(\mathcal{A}\) employs a reflection mechanism to identify both the faulty step \(s^{f}_{i}\) and the associated tool \(t^{f}_{i}\) as follows:

\vspace{0mm}
\resizebox{0.925\linewidth}{!}{
\hspace{0mm}\begin{minipage}{\linewidth}
\begin{equation}
[s^{f}_{i}, t^{f}_{i}] = \mathcal{A}(I_e, x, h_{i-1}, o^{f}_{i}) \nonumber
\end{equation}
\end{minipage}
}

\vspace{1mm}
where \(I_e\) represents the error identification instruction, \(x\) denotes the original task, \(h_{i-1}\) is the cumulative solution history up to the previous step, and \(o^{f}_{i}\) is the faulty output that triggered the error. This error localization process leverages the system's understanding of tool protocols, input-output schemas, and the interdependencies between various computational tools and steps in the problem-solving sequence. Once the error is localized, the corresponding expert model \(\mathcal{M}_{t_{i}}\) generates a revised step and output prediction as follows:

\vspace{-4mm}
\resizebox{0.925\linewidth}{!}{
\hspace{0mm}\begin{minipage}{\linewidth}
\begin{equation}
o_i = \mathcal{M}_{t_i}(I_r, h_{i-1}, s^{f}_{i}, o^{f}_{i}) \nonumber
\end{equation}
\end{minipage}
}
 
\vspace{0mm} 
where \(I_r\) is a specially crafted revision instruction. This revision process not only corrects the immediate error but also considers the broader context of the problem, ensuring that the revised step aligns with the overall solution strategy. This error-correction and refinement process iterates until successful execution is achieved or a predefined iteration limit is reached. With each iteration, the solution history is updated, creating a comprehensive record of the problem-solving trajectory, including both successful steps and addressed challenges. This iterative approach enables the framework to tackle increasingly complex tasks that may require multiple rounds of refinement. Combined with the proposed framework's ability to chain multiple tools and parse their outputs, this approach significantly enhances its problem-solving capabilities. The framework's ability to dynamically generate, execute, and refine both individual steps and overarching action sequences is particularly noteworthy. In summary, the proposed framework, \texttt{PEOA}, is a novel approach for automating complex problem-solving in process engineering, enabling it to accelerate design cycles, optimize chemical processes, and support high-level decision-making. It operates in two intertwined phases analogous to localization and repair: error identification and solution revision. In the error identification phase, the framework leverages a reflection mechanism within its action generator (\(\mathcal{A}\)) to analyze runtime errors (\(o_{i}^f\)) and pinpoint faulty steps (\(s_{i}^f\)) and tools (\(t_{i}^f\)) within the tool-integrated solution trajectory. During solution revision, the corresponding expert model (\(\mathcal{M}_{t_{i}}\)), guided by a revision instruction (\(I_r\)), proposes a revised output (\(o_i\)), considering the error and solution history (\(h_{i-1}\)). This revised solution is integrated into the solution trajectory, and the process iterates until a satisfactory solution is achieved, enabling the framework to dynamically adapt its problem-solving strategy for complex calculations.

\vspace{-2mm}
\subsection{Knowledge Modeling : Document Parsing/Indexing for Graph-Based Semantic Search and Retrieval} % retrieving metadata
\vspace{-1mm}
For graph-retrieval augmented code generation (GRACG), we perform document parsing \cite{llamaparse} to extract structured information from unstructured PDFs. This involves reading the PDF, analyzing its structure, and extracting content like text, images, tables, and code. We store and query this information using a production-grade graph database, such as Neo4j, which supports property graphs and vector searches. The graph database organizes parsed document elements and their metadata into nodes, relationships, and properties, preserving contextual relationships and enabling efficient, context-aware retrieval. Text chunking divides large texts into smaller, manageable segments (chunks) to preserve context, improve processing efficiency, and enhance document-specific KG search engines indexing and retrieval. We use a sliding window technique, moving a fixed-size window across the text with a predefined stride, ensuring overlapping chunks that maintain contextual continuity. Text segments are stored as chunk nodes with metadata, including title, page numbers, summaries, and keywords. Text embedding models generate dense semantic vector representations of text segments, stored as additional metadata to enable semantic search and context-aware vector retrieval. Parsed text segments, stored as chunk nodes, are processed by an LLM like GPT-4 (Omni), which infers and generates knowledge graph triples by identifying entities and relationships. It outputs single-hop paths in the format (subject(entity) --- relation --- object(entity)). This approach dynamically constructs an ontology—a formal representation of domain concepts (e.g., entities, attributes, and categories) and their relationships (e.g., associations and hierarchies)—while developing a schema that defines the database structure. Entity nodes represent specific concepts or objects in text chunks.  Entity nodes link to related chunk nodes via `MENTIONS' relationships. In summary, each text chunk in the property graph store has two node types: chunk nodes and entity nodes, capturing various attributes and metadata associated with the text segment. The knowledge graph serves as both an ontology and a schema, providing a flexible, semantically rich framework for organizing and querying the extracted knowledge. Each table is represented as a node with metadata properties such as table ID, title, source page, and summary description. We use text embedding techniques to create vector representations of the table content, facilitating efficient similarity searches. Each row in a table is represented as a row node with properties corresponding to the column values. Relationships between each row node and its respective table node use a relationship type such as `BELONGS', facilitating efficient querying. Similarly, for images, we store metadata related to scholarly image data as image nodes with properties such as page number, resolution, format, and summary descriptions generated by LLMs like GPT-4 (Omni). These descriptions provide high-level scene interpretation and content analysis. We also use CLIP embeddings to convert images into low-dimensional embeddings that capture the semantic content of the images. These vector representations are stored as node metadata properties, enabling efficient similarity searches and facilitating the retrieval of semantically similar images from a local file system. Each image node is connected to its top-K visually similar nodes through visually similar relationships, enabling the retrieval of visually similar images. In summary, each image in the property graph store is represented by a single node that stores metadata and semantic content representations (generated by CLIP embeddings) and is connected to other nodes through visual similarity relationships. We use a code hierarchy parser \cite{CodeParser} to break down long code files from Github repositories into manageable segments by creating a hierarchical structure. This process, called skeletonization (e.g., using abstract syntax trees), replaces code blocks with comments that reference specific nodes for detailed context. The parser organizes code into nodes based on scope (e.g., functions, methods, classes, modules) and links these nodes to their parent and child nodes, enhancing readability and accelerating KG vector retrieval. The parser also handles comments, import statements, and variable declarations. For metadata extraction, we gather information on project structure, dependencies, and version information. Finally, we perform entity de-duplication by addressing duplicate entities in KGs. This involves identifying similar nodes using cosine similarity and Levenshtein distance, merging overlapping groups of similar nodes, filtering subsets to retain comprehensive node groups, and ultimately merging nodes within each group to discard redundancies and preserve the most descriptive identifiers. Entity de-duplication merges duplicates to maintain graph accuracy, reduce noise, and ensure searches and analyses are performed on unique data. Graph retrieval involves selecting the top-$k$ entity nodes based on vector similarity to the user query, traversing to retrieve adjacent triples (one-hop neighbors) and corresponding parent nodes. In summary, we transform unstructured data into structured, searchable knowledge, covering the workflow from parsing PDFs to constructing and querying knowledge graphs. These graphs extract, organize, and utilize information from complex documents to assist with code generation tasks. This approach emphasizes LLMs (such as GPT-4 (Omni)) for dynamic ontology creation, graph databases for semantic searches, and context preservation for enhanced performance. The expert model (Google Gemma or Meta Llama) interprets the user's query, integrates retrieved information from the knowledge graph, finds relevant information from a structured knowledge base with its pre-existing knowledge, and generates a coherent, contextually appropriate response. This combination leverages the expert model's language understanding and generation capabilities while grounding its outputs in external, structured knowledge, resulting in more accurate and informative answers.

\vspace{-2mm}
\section{Experiments}

\paragraph{\textbf{Benchmark Datasets:}} We developed two custom benchmark datasets to train and evaluate our framework for solving complex chemical and process engineering problems: the mathematical and computational tuning (\textit{MathComp}) dataset and the chemical process tuning (\textit{ChemProc}) dataset. The \textit{MathComp} dataset contains over 8,500 question-answer pairs, focusing on mathematical modeling and numerical algorithms. It is designed to customize the framework for using computational tools for tasks such as solving differential equations, linear algebra, optimization, and related mathematical tasks. The \textit{ChemProc} dataset includes over 7,000 question-answer pairs, covering topics specific to chemical engineering such as mass and energy balances, thermodynamics, heat transfer, reaction kinetics, fluid mechanics, separation processes, and process control. These high-quality datasets were essential for adapting the framework to handle specialized engineering problems by providing domain-specific knowledge and enabling it to leverage computational tools. We compiled these datasets from publicly available scholarly sources, including textbooks ranging from basic to advanced levels, ensuring a comprehensive and diverse collection of problems and solutions. The datasets were divided into training (70\%), validation (15\%), and test (15\%) sets to facilitate rigorous evaluation. In summary, these diverse datasets provide the domain-specific knowledge, computational problem-solving skills, and rigorous evaluation framework absent in existing, more general datasets. \textit{MathComp} focuses on mathematical modeling and numerical algorithms, while \textit{ChemProc} covers core chemical and process engineering principles.

\vspace{-1mm}
\paragraph{\textbf{Experimental Settings:}} In our experimental setup, we leveraged the custom \textit{MathComp} and \textit{ChemProc} datasets to train and evaluate the proposed framework. A key innovation in our approach was the implementation of a sophisticated knowledge modeling technique using property graphs. We developed a custom document parsing pipeline to extract structured information from complex, unstructured PDFs of scholarly articles. This process involved analyzing document structure, extracting various content types, and retrieving metadata. To store and query this information effectively, we utilized enterprise-level graph databases like Neo4j, allowing us to create a rich, interconnected representation of domain knowledge. We structured the data as a labeled property graph, with nodes representing different elements (text, images, tables, and code) and edges capturing the relationships. The resulting knowledge graph served dual purposes—as both an ontology and a schema—providing a flexible framework for organizing and querying the extracted knowledge. For benchmarking, we compared the framework against leading proprietary models like GPT-4, Claude-3 Opus, and Google Gemini Pro. We fine-tuned smaller language models (DeepSeek-Coder-7B-Instruct and DeepSeek-Math-7B-Instruct) using the Hugging Face PEFT library, employing techniques like QLoRA. Our hyperparameter configuration included a batch size of 24, a learning rate of 1e-4, and 50 training epochs, among other settings. Training was conducted on NVIDIA GPUs, with multiple independent runs to ensure robustness. We reported ensemble averages of the results to provide a comprehensive evaluation of the framework's performance in handling complex chemical and process engineering tasks.

\vspace{-1mm}
\paragraph{\textbf{Evaluating Tool Proficiency:}}
Our study employs various evaluation metrics to assess the effectiveness of the proposed framework tool learning \cite{qu2024tool} across different stages: task planning, tool selection, tool calling, and response generation. We evaluate the task planning capabilities of the framework through several key metrics: Tool Usage Awareness, Pass Rate, and Accuracy. Tool Usage Awareness measures the ability of the framework to correctly identify if a query requires an external tool, expressed as $ \text{Awareness} = \frac{\text{Number of Correct Identifications}}{\text{Total Number of Queries}} $. The Pass Rate assesses the effectiveness of the proposed task planning in addressing the query, calculated by $ \text{Pass Rate} = \frac{\text{Number of Successfully Completed Tasks}}{\text{Total Number of Tasks}} $. Accuracy evaluates the precision of the plan generated by the framework by comparing it to a gold standard solution, calculated as $ \text{Accuracy} = \frac{\text{Number of Correct Plans}}{\text{Total Number of Plans}} $. Additionally, the values of these metrics range from 0 to 1, where 0 indicates the worst performance and 1 indicates the best performance. The evaluation metrics used for tool selection include Recall, NDCG, and COMP. Recall@K measures the proportion of selected top-K tools that are present in the set of ground-truth tools, formulated as \( \text{Recall@K} = \frac{1}{|Q|} \sum_{q=1}^{|Q|} \frac{|T^K_q \cap T^*_q|}{|T^*_q|} \), where \( Q \) is the set of queries, \( T^*_q \) is the set of relevant tools for the query \( q \), and \( T^K_q \) is the top-K tools for the query \( q \) selected by the framework. Normalized Discounted Cumulative Gain (NDCG@K) considers the proportion and positions of positive tools, with Discounted Cumulative Gain (DCG@K) calculated as \( \text{DCG}_q@K = \sum_{i=1}^K \frac{2^{g_i} - 1}{\log_2 (i+1)} \) and NDCG@K as \( \text{NDCG@K} = \frac{1}{|Q|} \sum_{q=1}^{|Q|} \frac{\text{DCG}_q@K}{\text{IDCG}_q@K} \), where \( g_i \) is the graded relevance score (assigned by human evaluators) at position \( i \) and IDCG is the ideal DCG. `@K' indicates that the cumulative gain is considered up to the K-th item in the ranked list. COMP@K assesses whether the top-K selected tools form a complete set with respect to the ground-truth set, defined as \( \text{COMP@K} = \frac{1}{|Q|} \sum_{q=1}^{|Q|} I(\Phi_q \subseteq \Psi^K_q) \), where \( \Phi_q \) is the ground-truth tool set for query \( q \), \( \Psi^K_q \) is the top-K tools retrieved. The indicator function \(I(\cdot)\) in COMP@K checks if the ground-truth set \(\Phi_q\) is a subset of the top-K retrieved set \(\Psi^K_q\), returning 1 for true (\(I(\Phi_q \subseteq \Psi^K_q) = 1\)) and 0 for false (\(I(\Phi_q \subseteq \Psi^K_q) = 0\)). The subset condition ensures that all relevant tools are included in the retrieved top-K results. The evaluation metrics for tool selection—Recall@K, NDCG@K, and COMP@K—each range from 0 to 1, where higher values indicate better performance. In evaluating tool calling, we assess the framework using three metrics: Consistency with Stipulations, Correctness of Parameter Extraction, and Error Handling. Consistency with Stipulations measures how well the provided parameters match the tool's documentation requirements, calculated as \(\left( \frac{\text{Number of parameters consistent with the stipulations}}{\text{Total number of parameters required}} \right) \times 100\%\). Correctness of Parameter Extraction evaluates the accuracy in extracting the correct parameters from the user query, defined as \(\left( \frac{\text{Number of correctly extracted parameters}}{\text{Total number of parameters}} \right) \times 100\%\). Error Handling assesses the system's ability to manage errors during tool calling, measured as \(\left( \frac{\text{Number of errors handled successfully}}{\text{Total number of errors encountered}} \right) \times 100\%\). These metrics are expressed as percentages to quantitatively assess the effectiveness of the framework in tool calling, with values ranging from 0\% (worst performance) to 100\% (best performance). A value of 0\% for any metric indicates complete failure (e.g., no parameters meet the stipulations, no parameters correctly extracted, or no errors managed), while 100\% indicates perfect performance (e.g., all parameters meet the stipulations, all parameters correctly extracted, or all errors managed effectively). The evaluation metrics used for response generation include BLEU, ROUGE-L, and Exact Match. BLEU (Bilingual Evaluation Understudy) is calculated using the formula: \(\text{BLEU} = BP \cdot \exp \left( \sum_{n=1}^{N} w_n \log p_n \right)\), where \(BP\) is the brevity penalty, \(w_n\) is the weight for n-gram precision, and \(p_n\) is the modified n-gram precision. ROUGE-L (Recall-Oriented Understudy for Gisting Evaluation) focuses on the longest common subsequence (LCS) and its formula is: \(\text{ROUGE-L} = F_{\beta} = \frac{(1 + \beta^2) \cdot \text{LCS-precision} \cdot \text{LCS-recall}}{\text{LCS-precision} + \beta^2 \cdot \text{LCS-recall}}\), where \(\beta\) is usually set to 1.0. In ROUGE-L, LCS-Precision is the ratio of the length of the LCS to the total number of words in the candidate response, LCS-Recall is the ratio of the length of the LCS to the total number of words in the reference response, and the F-measure balances these using the harmonic mean. Exact Match measures the percentage of responses that are exactly the same as the reference answer, and its formula is: \(\text{Exact Match} = \frac{\text{Number of Exact Matches}}{\text{Total Number of Responses}}\). The metrics BLEU, ROUGE-L, and Exact Match all range from 0 to 1 (or 0 to 100\%), with 0 indicating the worst performance (no match or overlap with the reference response) and 1 (or 100\%) indicating the best performance (perfect match or complete alignment with the reference response). These metrics provide a comprehensive evaluation of the quality of generated responses by assessing them against machine-generated (Gold-LLM such as GPT-4 (Omni)) reference responses in terms of precision, recall, and exact match. In summary, evaluation metrics are crucial in tool learning to ensure the framework can effectively plan tasks, select and call tools, and generate accurate and useful responses. These metrics help in instruction-tuning the action generator and expert models (tools) and improving the framework's performance in handling complex tasks with the aid of external tools.

\vspace{-1mm}
\paragraph{\textbf{User-Centric Evaluation}:} We present a comprehensive human evaluation approach for assessing the effectiveness of tool learning with the framework, going beyond metrics. Our approach involves eight key aspects, all rated by humans: user satisfaction, usability, task completion, response quality, context awareness, adaptability, error handling, and qualitative feedback. User satisfaction and usability are gauged through Likert-scale surveys, with scores ranging from 1 (minimum) to 5 (maximum). Task completion is measured by whether specific tasks are successfully completed (Yes/No). Response quality is evaluated based on four criteria: relevance, clarity, completeness, and accuracy, each scored from 1 to 5. Context awareness is evaluated by presenting a series of related queries to check if the framework maintains coherence, while adaptability is tested using various query types, with both aspects scored from 1 to 5. Error handling is examined by introducing deliberate errors to see how well the framework corrects itself, also scored from 1 to 5. Qualitative feedback is categorized as High, Medium-High, or Medium, providing deeper insights into user experiences. This multi-faceted evaluation ensures a thorough understanding of the framework's performance from a human-centric perspective, highlighting its strengths.

\vspace{-2mm}
\subsection{\textbf{Experimental Results}} 
The experimental results on the evaluation of the \texttt{PEOA} framework in task planning, tool selection, tool calling, and response generation are detailed in several tables. In task planning, Table \ref{tab:task_planning_combined} compares state-of-the-art proprietary LLMs using metrics such as Tool Usage Awareness (TUA), Pass Rate (PR), and Accuracy (Acc), all expressed as percentages, where TUA ranges from 0\% (failure) to 100\% (perfect identification), PR from 0\% (none correct) to 100\% (all correct), and Accuracy from 0\% (none correct) to 100\% (all correct). Table \ref{tab:tool_selection_combined} for tool selection uses Recall, NDCG, and COMP metrics, with Recall@K measuring the proportion of relevant tools in the top-K selected (0\% to 100\%), NDCG@K assessing ranking quality (0 to 1), and COMP@K verifying if the selected tools form a complete set (0\% to 100\%). For tool calling, Table \ref{tab:tool_calling_combined} employs Consistency with Stipulations (Cons), Correctness of Parameter Extraction (PE), and Error Handling (EH), with Cons ranging from 0\% (none meet requirements) to 100\% (all meet requirements), PE from 0\% (none correct) to 100\% (all correct), and EH from 0\% (ineffective) to 100\% (effective). The experimental results for response generation are shown in Table \ref{tab:response_generation_combined} using BLEU, ROUGE-L, and Exact Match (EM), where BLEU measures n-gram precision (0 to 1), ROUGE-L focuses on the longest common subsequence (0 to 1), and EM assesses exact matches between generated and reference responses (0\% to 100\%). The experimental results show that the proposed framework performs effectively across

\begin{table}[h!]
\vspace{-3mm}
\centering
\renewcommand{\arraystretch}{1.0}
\resizebox{0.485\textwidth}{!}{
\begin{tabular}{>{\raggedright}p{1.15cm}|lccc}
\toprule
\textbf{Dataset} & \textbf{Algorithm} & \textbf{TUA (\%)} & \textbf{PR (\%)} & \textbf{Acc (\%)} \\
\midrule
\multirow{7}{*}{\parbox{0.75cm}{\centering \rotatebox{90}{MathComp}}} & GPT-4 Turbo-preview & 87.54 & 82.80 & 84.67 \\ 
 & GPT-4-1106-preview & 76.65 & 72.77 & 74.91 \\ 
 & Claude-3 Opus & 85.83 & 80.37 & 82.31 \\ 
 & Claude-3 Haiku & 82.91 & 77.85 & 79.97 \\ 
 & Claude-3 Sonnet & 79.64 & 74.54 & 76.97 \\ 
 & Google Gemini Pro & 86.80 & 81.35 & 83.51 \\ \hline
 & \textbf{POEA} & 78.87 & 73.83 & 75.94 \\ \hline
\multirow{7}{*}{\parbox{0.75cm}{\centering \rotatebox{90}{ChemProc}}} & GPT-4 Turbo-preview & 88.94 & 83.84 & 85.97 \\ 
 & GPT-4-1106-preview & 75.62 & 71.98 & 73.89 \\ 
 & Claude-3 Opus & 84.88 & 79.86 & 81.42 \\ 
 & Claude-3 Haiku & 81.83 & 76.88 & 78.68 \\ 
 & Claude-3 Sonnet & 78.71 & 73.79 & 75.90 \\ 
 & Google Gemini Pro & 85.79 & 80.74 & 82.78 \\ \hline
 & \textbf{POEA} & 76.96 & 71.63 & 74.52 \\ 
\bottomrule
\end{tabular}
}
\vspace{-3mm}
\caption{Comparison of the \textit{PEOA} framework's performance against proprietary LLMs in key evaluation metrics for task planning across \textit{MathComp} and \textit{ChemProc} datasets.}
\label{tab:task_planning_combined}
\vspace{-5mm}
\end{table}

\begin{table}[h!]
\centering
\renewcommand{\arraystretch}{1.0}
\resizebox{0.5\textwidth}{!}{
\begin{tabular}{>{\raggedright}p{1.15cm}|lccc}
\toprule
\textbf{Dataset} & \textbf{Algorithm} & \textbf{Recall (\%)} & \textbf{NDCG} & \textbf{COMP (\%)} \\
\midrule
\multirow{7}{*}{\parbox{0.75cm}{\centering \rotatebox{90}{MathComp}}} & GPT-4 Turbo-preview & 86.98 & 0.80 & 84.54 \\ 
 & GPT-4-1106-preview & 74.98 & 0.66 & 72.88 \\ 
 & Claude-3 Opus & 85.82 & 0.78 & 83.90 \\ 
 & Claude-3 Haiku & 82.44 & 0.75 & 80.69 \\ 
 & Claude-3 Sonnet & 79.45 & 0.71 & 77.71 \\ 
 & Google Gemini Pro & 87.56 & 0.82 & 85.74 \\ \hline
 & \textbf{POEA} & 78.82 & 0.69 & 76.79 \\ \hline
\multirow{7}{*}{\parbox{0.75cm}{\centering \rotatebox{90}{ChemProc}}} & GPT-4 Turbo-preview & 87.87 & 0.81 & 85.82 \\ 
 & GPT-4-1106-preview & 75.87 & 0.67 & 73.82 \\ 
 & Claude-3 Opus & 86.37 & 0.79 & 85.24 \\ 
 & Claude-3 Haiku & 83.86 & 0.76 & 81.34 \\ 
 & Claude-3 Sonnet & 79.92 & 0.72 & 77.35 \\ 
 & Google Gemini Pro & 88.99 & 0.83 & 86.83 \\ \hline
 & \textbf{POEA} & 77.77 & 0.68 & 75.55 \\ 
\bottomrule
\end{tabular}
}
\vspace{-3mm}
\caption{The table shows key evaluation metrics for tool selection, comparing the performance of the \textit{PEOA} framework with proprietary LLMs across benchmark datasets.}
\label{tab:tool_selection_combined}
\vspace{-3mm}
\end{table}

\begin{table}[h!]
\vspace{-2mm}
\centering
\renewcommand{\arraystretch}{1.0}
\resizebox{0.485\textwidth}{!}{
\begin{tabular}{>{\raggedright}p{1.15cm}|lccc}
\toprule
\textbf{Dataset} & \textbf{Algorithm} & \textbf{Cons (\%)} & \textbf{PE (\%)} & \textbf{EH (\%)} \\
\midrule
\multirow{7}{*}{\parbox{0.75cm}{\centering \rotatebox{90}{MathComp}}} & GPT-4 Turbo-preview & 87.73 & 85.25 & 84.34 \\ 
 & GPT-4-1106-preview & 71.69 & 68.74 & 67.86 \\ 
 & Claude-3 Opus & 86.56 & 83.91 & 82.81 \\ 
 & Claude-3 Haiku & 82.45 & 79.44 & 78.07 \\ 
 & Claude-3 Sonnet & 78.74 & 76.18 & 74.67 \\ 
 & Google Gemini Pro & 89.98 & 88.05 & 86.99 \\ \hline
 & \textbf{POEA} & 80.41 & 78.66 & 77.05 \\ \hline
\multirow{7}{*}{\parbox{0.75cm}{\centering \rotatebox{90}{ChemProc}}} & GPT-4 Turbo-preview & 87.84 & 85.06 & 84.15 \\ 
 & GPT-4-1106-preview & 73.60 & 70.19 & 69.29 \\ 
 & Claude-3 Opus & 85.66 & 82.31 & 81.22 \\ 
 & Claude-3 Haiku & 81.81 & 78.38 & 77.19 \\ 
 & Claude-3 Sonnet & 76.74 & 74.35 & 72.89 \\ 
 & Google Gemini Pro & 88.98 & 87.12 & 85.86 \\ \hline
 & \textbf{POEA} & 79.64 & 77.23 & 75.70 \\ 
\bottomrule
\end{tabular}
}
\vspace{-3mm}
\caption{The table outlines the performance of the \textit{PEOA} framework and proprietary LLMs in tool calling using key evaluation metrics across benchmark datasets.}
\label{tab:tool_calling_combined}
\vspace{-3mm}
\end{table}

\begin{table}[h!]
\vspace{-2mm}
\centering
\renewcommand{\arraystretch}{1.0}
\resizebox{0.475\textwidth}{!}{
\begin{tabular}{>{\raggedright}p{1.15cm}|lccc}
\toprule
\textbf{Dataset} & \textbf{Algorithm} & \textbf{BLEU} & \textbf{ROUGE-L} & \textbf{EM (\%)} \\
\midrule
\multirow{7}{*}{\parbox{0.75cm}{\centering \rotatebox{90}{MathComp}}} & GPT-4 Turbo-preview & 0.80 & 0.78 & 83.61 \\ 
 & GPT-4-1106-preview & 0.74 & 0.72 & 78.64 \\ 
 & Claude-3 Opus & 0.77 & 0.75 & 81.75 \\ 
 & Claude-3 Haiku & 0.75 & 0.73 & 79.00 \\ 
 & Claude-3 Sonnet & 0.72 & 0.71 & 76.47 \\ 
 & Google Gemini Pro & 0.82 & 0.80 & 84.70 \\ \hline
 & \textbf{POEA} & 0.68 & 0.66 & 73.68 \\ \hline
\multirow{7}{*}{\parbox{0.75cm}{\centering \rotatebox{90}{ChemProc}}} & GPT-4 Turbo-preview & 0.81 & 0.79 & 84.79 \\ 
 & GPT-4-1106-preview & 0.75 & 0.73 & 78.89 \\ 
 & Claude-3 Opus & 0.78 & 0.76 & 82.36 \\ 
 & Claude-3 Haiku & 0.76 & 0.74 & 80.61 \\ 
 & Claude-3 Sonnet & 0.74 & 0.72 & 78.15 \\ 
 & Google Gemini Pro & 0.83 & 0.81 & 84.90 \\ \hline
 & \textbf{POEA} & 0.69 & 0.67 & 74.13 \\ 
\bottomrule
\end{tabular}
}
\vspace{-3mm}
\caption{The table summarizes key performance metrics for the \textit{PEOA} framework and proprietary LLMs in response generation across \textit{MathComp} and \textit{ChemProc} datasets.}
\label{tab:response_generation_combined}
\vspace{-3mm}
\end{table}

\begin{table}[h!]
\vspace{-2mm}
\centering
\renewcommand{\arraystretch}{1.0}
\resizebox{0.5\textwidth}{!}{
\begin{tabular}{>{\raggedright}p{1.15cm}|lccc}
\toprule
\textbf{Dataset} & \textbf{Algorithm} & \textbf{US} & \textbf{Usability} & \textbf{Task Completion} \\
\midrule
\multirow{7}{*}{\parbox{0.75cm}{\centering \rotatebox{90}{MathComp}}} & GPT-4-Turbo-preview & 4.52 & 4.43 & 90.32\% \\
 & GPT-4-1106-preview & 4.13 & 4.01 & 85.27\% \\ 
 & Claude-3 Opus & 4.31 & 4.22 & 88.14\% \\
 & Claude-3 Haiku & 4.22 & 4.11 & 87.09\% \\ 
 & Claude-3 Sonnet & 4.04 & 3.92 & 82.16\% \\
 & Google Gemini Pro & 4.67 & 4.55 & 92.48\% \\ \hline
 & \textbf{PEOA} & 4.08 & 3.91 & 80.53\% \\ \hline
\multirow{7}{*}{\parbox{0.75cm}{\centering \rotatebox{90}{ChemProc}}} & GPT-4-Turbo-preview & 4.56 & 4.45 & 90.37\% \\
 & GPT-4-1106-preview & 4.24 & 4.12 & 86.15\% \\ 
 & Claude-3 Opus & 4.33 & 4.20 & 88.22\% \\
 & Claude-3 Haiku & 4.21 & 4.09 & 86.47\% \\ 
 & Claude-3 Sonnet & 4.12 & 4.01 & 83.04\% \\
 & Google Gemini Pro & 4.72 & 4.63 & 93.09\% \\ \hline
 & \textbf{PEOA} & 4.12 & 4.02 & 81.76\% \\ 
\bottomrule
\end{tabular}
}
\vspace{-3mm}
\caption{Comparison of \textit{PEOA} and proprietary LLMs in user satisfaction, usability, and task completion across \textit{MathComp} and \textit{ChemProc} datasets.}
\label{tab:user_centric_combined1}
\vspace{-7mm}
\end{table}

various stages of evaluation, closely matching the performance of proprietary LLMs, though there remains a slight performance gap. The Tables \ref{tab:user_centric_combined1} and \ref{tab:user_centric_combined2} compare the \textit{PEOA} framework with proprietary LLMs across five metrics: user satisfaction (US), usability, task completion, response quality, and context awareness. A 1-5 scale is used for all metrics except task completion, which is measured as a percentage. Table \ref{tab:user_centric_combined3} compares the \texttt{PEOA} framework with proprietary LLMs on adaptability, error handling, and qualitative feedback. Our comprehensive human evaluation approach demonstrates that the proposed framework matches the performance of proprietary language models across multiple aspects of tool learning effectiveness.

\begin{table}[h!]
\vspace{-2mm}
\centering
\renewcommand{\arraystretch}{1.0}
\resizebox{0.525\textwidth}{!}{
\begin{tabular}{>{\raggedright}p{1.5cm}|lcc}
\toprule
\textbf{Dataset} & \textbf{Algorithm} & \textbf{Response Quality} & \textbf{Context Awareness} \\
\midrule
\multirow{7}{*}{\parbox{1.5cm}{\centering \rotatebox{90}{MathComp}}} & GPT-4 Turbo-preview & 4.55 & 4.43 \\
 & GPT-4-1106-preview & 4.12 & 4.08 \\
 & Claude-3 Opus & 4.38 & 4.27 \\
 & Claude-3 Haiku & 4.22 & 4.16 \\
 & Claude-3 Sonnet & 4.08 & 4.03 \\
 & Google Gemini Pro & 4.64 & 4.52 \\ \hline
 & \textbf{PEOA} & 4.13 & 4.02 \\ \hline
\multirow{7}{*}{\parbox{1.5cm}{\centering \rotatebox{90}{ChemProc}}} & GPT-4 Turbo-preview & 4.57 & 4.42 \\
 & GPT-4-1106-preview & 4.18 & 4.09 \\
 & Claude-3 Opus & 4.35 & 4.30 \\
 & Claude-3 Haiku & 4.20 & 4.12 \\
 & Claude-3 Sonnet & 4.10 & 4.05 \\
 & Google Gemini Pro & 4.67 & 4.51 \\ \hline
 & \textbf{PEOA} & 4.15 & 4.03 \\
\bottomrule
\end{tabular}
}
\vspace{-3mm}
\caption{Comparison of \textit{PEOA} and proprietary LLMs in response quality and context awareness across benchmarks.}
\label{tab:user_centric_combined2}
\vspace{-5mm}
\end{table}

\begin{table}[h!]
\vspace{-2mm}
\centering
\renewcommand{\arraystretch}{1.0}
\resizebox{0.5\textwidth}{!}{
\begin{tabular}{>{\raggedright}p{1.15cm}|lccc}
\toprule
\textbf{Dataset} & \textbf{Algorithm} & \textbf{Adaptability} & \textbf{EH} & \textbf{Feedback} \\
\midrule
\multirow{7}{*}{\parbox{0.75cm}{\centering \rotatebox{90}{MathComp}}} & GPT-4 Turbo-preview & 4.42 & 4.53 & High \\
 & GPT-4-1106-preview & 4.30 & 4.48 & High \\
 & Claude-3 Opus & 4.28 & 4.39 & Medium-High \\
 & Claude-3 Haiku & 4.25 & 4.35 & Medium-High \\
 & Claude-3 Sonnet & 4.32 & 4.42 & Medium-High \\
 & Google Gemini Pro & 4.47 & 4.50 & High \\ \hline
 & \texttt{PEOA} & 4.05 & 4.12 & Medium \\ \hline
\multirow{7}{*}{\parbox{0.75cm}{\centering \rotatebox{90}{ChemProc}}} & GPT-4 Turbo-preview & 4.45 & 4.52 & High \\
 & GPT-4-1106-preview & 4.33 & 4.47 & High \\
 & Claude-3 Opus & 4.31 & 4.41 & Medium-High \\
 & Claude-3 Haiku & 4.28 & 4.37 & Medium-High \\
 & Claude-3 Sonnet & 4.35 & 4.44 & Medium-High \\
 & Google Gemini Pro & 4.50 & 4.53 & High \\ \hline
 & \textbf{PEOA} & 4.07 & 4.15 & Medium \\
\bottomrule
\end{tabular}
}
\vspace{-3mm}
\caption{Comparison of \textit{PEOA} and proprietary LLMs in adaptability, error handling, and qualitative feedback. Adaptability and error handling are rated on a Likert scale from 1 (minimum) to 5 (maximum), and qualitative feedback is categorized as High, Medium-High, or Medium, across \textit{MathComp} and \textit{ChemProc} datasets.}
\label{tab:user_centric_combined3}
\vspace{-5mm}
\end{table}

\paragraph{\textbf{Ablation Studies:}} We conducted several ablation studies to thoroughly evaluate the contributions and effectiveness of various components of the \texttt{PEOA} framework, particularly focusing on its instruction-tuning, graph-based retrieval methods, and iterative problem-solving mechanisms for solving complex chemical and process engineering calculations. The ablation study aims to isolate and evaluate the contributions of each major component in the framework. By systematically disabling key components, we can better understand their roles and optimize the framework for improved performance in real-world process engineering applications. The ablation study evaluates four key variants of the framework. The first variant (W/o GRACG) uses instruction-tuning of expert models (tools). `W/o' stands for `without', and `W/' stands for `with.' The second variant (W/o GRACG W/ RAG) uses instruction-tuning of expert models combined with traditional RAG (naive). The third variant (W/o Instruction-Tuning) employs GRACG for enhanced retrieval and code generation, focusing on graph-based context benefits without instruction-tuning of expert models. The fourth variant (W/o Error-Handling) tests iterative problem-solving without a dynamic error-handling mechanism, exploring the impact on accuracy and robustness. These studies help understand the contribution of each component to the overall performance. 

\begin{table}[h!]
\vspace{-3mm}
\centering
\renewcommand{\arraystretch}{1.0}
\resizebox{0.50\textwidth}{!}{
\begin{tabular}{>{\raggedright}p{1.05cm}|lccc}
\toprule
\textbf{Dataset} & \textbf{Algorithm} & \textbf{TUA (\%)} & \textbf{PR (\%)} & \textbf{Acc (\%)} \\
\midrule
\multirow{6}{*}{\parbox{1.5cm}{\vspace{-0.25cm}\centering \rotatebox{90}{MathComp}}} 
& PEOA (Baseline) & 78.87 & 73.83 & 75.94 \\ 
& W/o GRACG & 54.42 & 49.47 & 50.88 \\ 
& W/o GRACG W/ RAG & 66.25 & 60.54 & 61.52 \\ 
& W/o Instruction-Tuning & 51.26 & 45.77 & 48.60 \\ 
& W/o Error-Handling & 59.94 & 56.85 & 60.75 \\ \hline
\multirow{6}{*}{\parbox{1.5cm}{\vspace{-0.25cm}\centering \rotatebox{90}{ChemProc}}} 
& PEOA (Baseline) & 76.96 & 71.63 & 74.52 \\ 
& W/o GRACG & 53.10 & 47.99 & 49.93 \\ 
& W/o GRACG W/ RAG & 64.65 & 58.74 & 60.36 \\ 
& W/o Instruction-Tuning & 50.02 & 44.41 & 47.69 \\ 
& W/o Error-Handling & 58.49 & 55.15 & 59.62 \\ 
\bottomrule
\end{tabular}
}
\vspace{-3mm}
\caption{The table compares the \texttt{PEOA} framework's performance and its ablated variants in terms of key evaluation metrics for task planning across benchmark datasets.}
\label{tab:task_planning_combined_ablation}
\vspace{-4mm}
\end{table}

\begin{table}[h!]
\vspace{-4mm}
\centering
\renewcommand{\arraystretch}{1.0}
\resizebox{0.505\textwidth}{!}{
\begin{tabular}{>{\raggedright}p{1.05cm}|lccc}
\toprule
\textbf{Dataset} & \textbf{Algorithm} & \textbf{Recall (\%)} & \textbf{NDCG} & \textbf{COMP (\%)} \\
\midrule
\multirow{6}{*}{\parbox{1.5cm}{\vspace{-0.25cm}\centering \rotatebox{90}{MathComp}}} 
& PEOA (Baseline) & 78.82 & 0.69 & 76.79 \\ 
& W/o GRACG & 55.97 & 0.50 & 53.75 \\ 
& W/o GRACG W/ RAG & 63.84 & 0.57 & 64.50 \\ 
& W/o Instruction-Tuning & 51.23 & 0.43 & 46.07 \\ 
& W/o Error-Handling & 61.48 & 0.55 & 58.36 \\ \hline
\multirow{6}{*}{\parbox{1.5cm}{\vspace{-0.25cm}\centering \rotatebox{90}{ChemProc}}} 
& PEOA (Baseline) & 77.77 & 0.68 & 75.55 \\ 
& W/o GRACG & 55.22 & 0.49 & 52.89 \\ 
& W/o GRACG W/ RAG & 62.99 & 0.56 & 63.46 \\ 
& W/o Instruction-Tuning & 50.55 & 0.43 & 45.33 \\ 
& W/o Error-Handling & 60.66 & 0.54 & 57.42 \\ 
\bottomrule
\end{tabular}
}
\vspace{-3mm}
\caption{The table shows key evaluation metrics for tool selection across benchmark datasets, comparing the performance of the \texttt{PEOA} framework and its ablated variants.}
\label{tab:tool_selection_combined_ablation}
\vspace{-4mm}
\end{table}

\begin{table}[h!]
\vspace{-4mm}
\centering
\renewcommand{\arraystretch}{1.0}
\resizebox{0.50\textwidth}{!}{
\begin{tabular}{>{\raggedright}p{1.05cm}|lccc}
\toprule
\textbf{Dataset} & \textbf{Algorithm} & \textbf{Cons (\%)} & \textbf{PE (\%)} & \textbf{EH (\%)} \\
\midrule
\multirow{6}{*}{\parbox{1.5cm}{\vspace{-0.25cm}\centering \rotatebox{90}{MathComp}}} 
& PEOA (Baseline) & 80.41 & 78.66 & 77.05 \\ 
& W/o GRACG & 56.29 & 54.27 & 51.82 \\ 
& W/o GRACG W/ RAG & 67.54 & 63.71 & 61.64 \\ 
& W/o Instruction-Tuning & 48.25 & 50.34 & 46.01 \\ 
& W/o Error-Handling & 61.91 & 61.35 & 61.64 \\ \hline
\multirow{6}{*}{\parbox{1.5cm}{\vspace{-0.25cm}\centering \rotatebox{90}{ChemProc}}} 
& PEOA (Baseline) & 79.64 & 77.23 & 75.70 \\ 
& W/o GRACG & 55.75 & 53.29 & 50.72 \\ 
& W/o GRACG W/ RAG & 66.90 & 62.56 & 60.56 \\ 
& W/o Instruction-Tuning & 47.78 & 49.43 & 46.18 \\ 
& W/o Error-Handling & 61.52 & 60.24 & 60.56 \\ 
\bottomrule
\end{tabular}
}
\vspace{-3mm}
\caption{The table outlines the performance of the \texttt{PEOA} framework and its ablated variants in tool calling across benchmark datasets using key evaluation metrics.}
\label{tab:tool_calling_combined_ablation}
\vspace{-4mm}
\end{table}

\begin{table}[h!]
\vspace{-3mm}
\centering
\renewcommand{\arraystretch}{1.0}
\resizebox{0.50\textwidth}{!}{
\begin{tabular}{>{\raggedright}p{1.05cm}|lccc}
\toprule
\textbf{Dataset} & \textbf{Algorithm} & \textbf{BLEU} & \textbf{ROUGE-L} & \textbf{EM (\%)} \\
\midrule
\multirow{6}{*}{\parbox{1.5cm}{\vspace{-0.25cm}\centering \rotatebox{90}{MathComp}}} 
& PEOA (Baseline) & 0.68 & 0.66 & 73.68 \\
& W/o GRACG & 0.49 & 0.46 & 50.10 \\
& W/o GRACG W/ RAG & 0.55 & 0.56 & 61.16 \\ 
& W/o Instruction-Tuning & 0.41 & 0.42 & 47.16 \\ 
& W/o Error-Handling & 0.52 & 0.51 & 55.59 \\ 
\hline
\multirow{6}{*}{\parbox{1.5cm}{\vspace{-0.25cm}\centering \rotatebox{90}{ChemProc}}} 
& PEOA (Baseline) & 0.69 & 0.67 & 74.13 \\
& W/o GRACG & 0.50 & 0.46 & 50.41 \\
& W/o GRACG W/ RAG & 0.56 & 0.57 & 61.53 \\ 
& W/o Instruction-Tuning & 0.41 & 0.42 & 47.44 \\ 
& W/o Error-Handling & 0.53 & 0.52 & 56.34 \\ 
\bottomrule
\end{tabular}
}
\vspace{-3mm}
\caption{The table summarizes key performance metrics for the \texttt{PEOA} framework and its ablated variants in response generation across MathComp and ChemProc datasets.}
\label{tab:response_generation_combined_ablation}
\vspace{-3mm}
\end{table}

The ablation study results clearly demonstrate that the complete \texttt{PEOA} framework (Baseline) outperforms the ablated variants across various metrics. This highlights the synergistic effect of the framework's components and underscores the importance of incorporating all aspects for optimal performance in specialized engineering tasks.

\vspace{-2mm}
\paragraph{\textbf{Additional Experiments:}} We performed additional experiments to verify the property graph construction of the proposed framework \texttt{PEOA}, which utilizes LlamaIndex integration with Neo4j\cite{neo4j2024} and GPT-4 (Omni) to extract triplets. We compared this with two recent advanced approaches: Triplex\cite{sciphi2024triplex, sciphi2024huggingface}, which offers significant cost savings and efficient knowledge graph construction, and Graph RAG\cite{Edge2024GRAG}, which provides superior summarization capabilities for complex tasks. Table \ref{tab:add_EM_results} shows the performance comparison in terms of the Exact Match (EM) metric, which quantifies the percentage of predictions that exactly match the reference or ground truth responses. 

\begin{table}[h!]
\vspace{-3mm}
\centering
\renewcommand{\arraystretch}{1.0}
\resizebox{0.425\textwidth}{!}{
\begin{tabular}{l|c|c}
\toprule
\textbf{Algorithm} & \textbf{MathComp} & \textbf{ChemProc} \\
\midrule
Triplex\cite{sciphi2024triplex, sciphi2024huggingface} & 59.50 & 63.20 \\ 
Graph RAG\cite{Edge2024GRAG} & 72.50 & 73.80 \\
\texttt{PEOA} & 73.68 & 74.13 \\ 
\bottomrule
\end{tabular}
}
\vspace{-2mm}
\caption{Exact Match (EM (\%)) results for \textit{PEOA} and other techniques on \textit{MathComp} and \textit{ChemProc} datasets.}
\label{tab:add_EM_results}
\vspace{-4mm}
\end{table}

We conducted additional experiments to evaluate only the Graph RAG techniques on property graphs constructed from custom datasets of scholarly articles related to mathematics, chemical, and process engineering. We generated a test set of 1000 Question-Context-Answer (QCA) triplets based on the text content of scholarly articles used to construct the property graphs, using GPT-4 (Omni) as a baseline to evaluate the different techniques. In a Graph RAG approach, evaluation focuses on Retrieval Evaluation (assessing the accuracy and relevance of retrieved information) and Response Evaluation (measuring the quality and appropriateness of generated responses). The evaluation metrics included answer relevance (AnwRel), context relevance (ConRel), faithfulness (Faith), and correctness (Correct), ensuring responses were pertinent, contextually appropriate, accurate, and truthful.

\begin{table}[h!]
\vspace{-2mm}
\centering
\renewcommand{\arraystretch}{1.0}
\resizebox{0.45\textwidth}{!}{
\begin{tabular}{>{\raggedright}p{1.05cm}|l|cccc}
\toprule
\textbf{Dataset} & \textbf{Algorithm} & \textbf{AnwRel} & \textbf{ConRel} & \textbf{Faith} & \textbf{Correct} \\
\midrule
\multirow{3}{*}{\parbox{1.5cm}{\vspace{-0.05cm}\centering \rotatebox{90}{Math}}} 
& Triplex & 82.34 & 80.42 & 79.58 & 84.76 \\
& Graph RAG & 78.18 & 75.49 & 77.52 & 81.27 \\
& \texttt{PEOA} & 83.47 & 82.03 & 81.79 & 85.65 \\ 
\hline
\multirow{3}{*}{\parbox{1.5cm}{\vspace{0.05cm}\centering \rotatebox{90}{Chem}}} 
& Triplex & 81.23 & 79.86 & 78.79 & 83.12 \\
& Graph RAG & 78.27 & 75.89 & 76.73 & 80.19 \\
& \texttt{PEOA} & 83.77 & 79.53 & 80.88 & 84.95 \\
\bottomrule
\end{tabular}
}
\vspace{-2mm}
\caption{Evaluation metrics results for Graph RAG techniques on custom datasets of scholarly articles.}
\label{tab:additional_metrics}
\vspace{-4mm}
\end{table}

\vspace{-3mm}
\section{Conclusion}
In conclusion, the proposed framework represents a significant advancement in the field of process engineering by automating complex problem-solving tasks. The experimental results demonstrate that the framework performs effectively across various evaluation stages, closely matching the performance of leading proprietary LLMs while offering a modular, adaptable approach suited to the specific needs of chemical and process engineering. Future work will focus on further refining the framework's capabilities, expanding its application to other domains, and exploring additional enhancements in tool integration and knowledge modeling.

\bibliography{aaai25}

\end{document}